\documentclass[letterpaper,10pt,conference]{ieeeconf}
\IEEEoverridecommandlockouts
\usepackage[T1]{fontenc}
\usepackage{marvosym}
\usepackage{amsmath,amssymb,bm}
\usepackage{graphicx,booktabs,tabularx,array,multirow}
\usepackage[table]{xcolor}
\usepackage{tikz}
\usetikzlibrary{arrows.meta,positioning,calc,fit,backgrounds}
\usepackage{cite,url,placeins}
\usepackage[ruled,lined,linesnumbered]{algorithm2e}
\SetKwInput{KwInput}{Input}
\SetKwInput{KwOutput}{Output}
\DontPrintSemicolon
\SetAlFnt{\small}
\SetAlCapFnt{\small\bfseries}
\SetAlCapNameFnt{\small}
\SetAlgoCaptionSeparator{:}
\SetAlgoNlRelativeSize{-1}
\SetCommentSty{textit}
\usepackage[labelfont=bf]{caption}
\usepackage{fancyhdr}
\usepackage[hidelinks]{hyperref}
\hypersetup{pdftitle={CoRe-WAM: Correspondence-Aligned Temporal Residuals for World Action Models},pdfauthor={Bin Zhou, Jialong Liu, Jianan Wang, Changhao Chen, Kani Chen},pdfsubject={Research preprint},pdfkeywords={world-action models, temporal correspondence, residual adaptation, robot manipulation}}
\definecolor{coreblue}{HTML}{2E7CC4}
\definecolor{corepurple}{HTML}{8E5AC8}
\definecolor{coreorange}{HTML}{D96B2B}
\definecolor{corebluefill}{HTML}{EDF6FF}
\definecolor{corepurplefill}{HTML}{F7F1FF}
\definecolor{coreorangefill}{HTML}{FFF3E8}
\definecolor{coreink}{HTML}{23344D}
\colorlet{traceblue}{coreblue}
\colorlet{studentorange}{coreorange}

\definecolor{gaincolor}{HTML}{18734B}
\definecolor{gapcolor}{HTML}{B74747}
\newcommand{\gain}[1]{\textcolor{gaincolor}{\textbf{#1}}}
\newcommand{\gap}[1]{\textcolor{gapcolor}{\scriptsize (#1)}}
\newcommand{\yes}{\ensuremath{\checkmark}}
\newcommand{\no}{\ensuremath{\times}}
\newcommand{\method}{CoRe-WAM}

\newcommand{\R}{\mathbb{R}}
\newcommand{\traceClean}{92.22}
\newcommand{\traceRand}{89.60}
\newcommand{\traceAvg}{90.91}
\newcommand{\traceCleanDelta}{+3.56}
\newcommand{\traceRandDelta}{+2.58}
\newcommand{\rawClean}{90.16}
\newcommand{\rawRand}{87.66}
\newcommand{\rawAvg}{88.91}
\newcommand{\rawCleanDelta}{+1.50}
\newcommand{\rawRandDelta}{+0.64}

\title{\LARGE\bf \raisebox{-5pt}{\includegraphics[height=24pt]{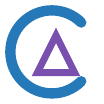}}\enspace CoRe-WAM: Correspondence-Aligned Temporal Residuals\\for World Action Models}
\author{Bin Zhou$^{1,3,\dagger}$, Jialong Liu$^{2,3,\dagger}$, Jianan Wang$^{3,\text{\Letter}}$, Changhao Chen$^{2,1,\text{\Letter}}$, and Kani Chen$^{1,\text{\Letter}}$
\thanks{$^1$The Hong Kong University of Science and Technology, Hong Kong SAR, China.}
\thanks{$^2$The Hong Kong University of Science and Technology (Guangzhou), Guangzhou, Guangdong, China.}
\thanks{$^3$Astribot, China.}
\thanks{$^{\dagger}$Bin Zhou and Jialong Liu contributed equally and share first authorship.}
\thanks{$^{\text{\Letter}}$Corresponding authors: Jianan Wang, Changhao Chen, and Kani Chen.}}
\begin{document}
\raggedbottom
\maketitle
\thispagestyle{fancy}
\pagestyle{fancy}
\begin{abstract}
Comparing current and past observations helps robots
understand scene changes and select subsequent actions
during manipulation. However, comparing visual features
at the same image location can mix different scene content
when objects or the camera move. We introduce CoRe-WAM,
a world-action model that incorporates correspondence-aligned
visual changes through a parameter-efficient temporal interface.
Its TraceDelta module uses correspondences from a frozen
tracking model to transport historical visual features to
current locations before computing signed differences in a
shared pretrained feature space.
Correspondence thus determines which historical content is
compared with the present, rather than entering the policy
as a separate trajectory representation.
A lightweight adapter converts these differences into
validity-gated residuals that supplement current visual
conditioning, allowing the policy to use recent changes
alongside current-scene information.
Built on Motus, CoRe-WAM keeps the pretrained backbone
weights frozen and optimizes 1.59 million parameters.
With a 5,000-update adaptation budget, CoRe-WAM achieves
\traceClean\% clean success across 50 RoboTwin 2.0 tasks, 3.56 percentage points above Motus;
on randomized evaluation, it achieves \traceRand\% success, a
2.58-point gain.
Integrating TraceDelta into a StarVLA-based policy improves
clean success from 58.10\% to 67.62\%, supporting transfer
of the temporal interface beyond Motus.
\end{abstract}
\suppressfloats[t]
\begin{figure}[t]
\centering
\includegraphics[width=\columnwidth,trim=10bp 13bp 11bp 9bp,clip]{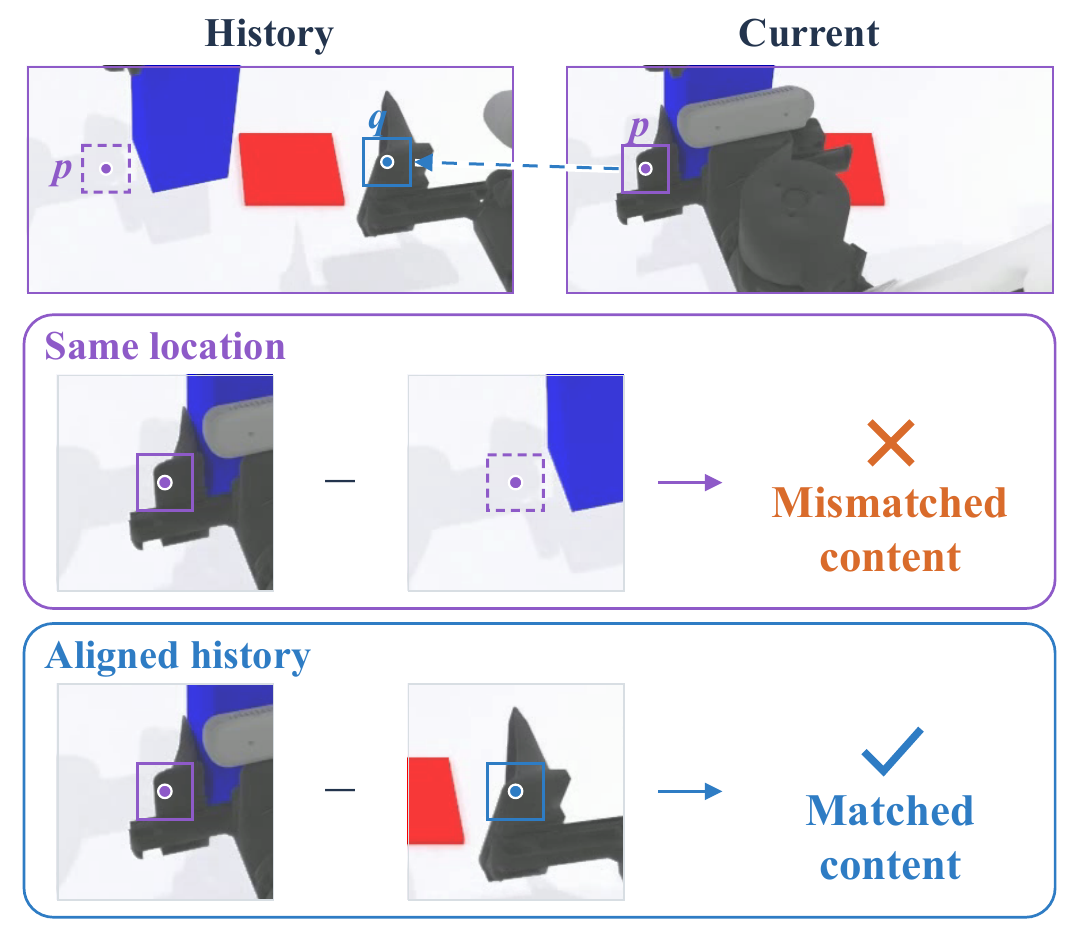}
\caption{\textbf{Why align history before differencing?}
The current point $p$ and historical point $p'$ share the
same image coordinate but correspond to different scene
content. Correspondence instead links $p$ to its historical
counterpart $q$, enabling feature differencing between
matched regions. Images and correspondences are illustrative.}
\label{fig:teaser}
\end{figure}
\section{Introduction}
\label{sec:introduction}
Robot manipulation changes where objects are and how they interact. A current observation describes the present scene, while recent observations reveal how that scene evolved through interaction. Across tool use, object handover, articulated-object interaction, and multi-stage placement, this observed evolution provides context for selecting subsequent actions. Vision-language-action models connect pretrained visual and language representations to robot control~\cite{brohan2023rt2,kim2024openvla,black2024pi0}. World-action models further connect control with visual prediction: Motus combines understanding, video-generation, and action experts~\cite{bi2025motus}, and Fast-WAM studies the role of video co-training and future imagination~\cite{yuan2026fastwam}. This leaves a basic question: how should a policy relate current and historical visual evidence so that observed changes can inform action generation?

Historical observations can supply more than additional scene appearances. CoMo learns continuous latent motion from videos for policy learning~\cite{yang2026como}, while trajectory-aware approaches expose temporal structure through predicted trajectories, motion features, or representation supervision~\cite{wen2024atm,yuan2026motionvla,yang2026fourDwam}. These approaches motivate using temporal evidence, but the choice of what to compare across observations determines how change is represented. A direct difference of pretrained features compares the same grid location in two frames. When an object or the camera moves, that location need not describe the same scene content (Fig.~\ref{fig:teaser}). The resulting difference mixes the evolution of a region's features with a change in which content occupies that region. Supplying history therefore does not, by itself, define a meaningful temporal comparison.

Our central idea is to establish correspondence before computing change. For each current region, we first find its estimated historical counterpart and then compare the two representations in the same pretrained feature space. This principle is related to registered-feature comparison in change detection, including CYWS-3D~\cite{sachdeva2023cyws3d}; here, we use it to construct robot-policy conditioning from observed interaction history. Unlike retrieving trajectory-field features as motion tokens~\cite{yuan2026motionvla} or aligning policy representations to trajectory-derived targets~\cite{yang2026fourDwam}, we use dense correspondence as a transport operator over pretrained visual features. The trajectory estimator determines \emph{where} to sample history, while the visual encoder determines \emph{what} is compared. The resulting aligned difference organizes temporal evidence around corresponding scene content rather than representing the trajectory itself.

Making this comparison useful for action generation imposes three requirements. First, temporal updates must be restricted to finite correspondences within the selected camera region; invalid sampling locations should contribute nothing. Second, differences in visual-feature space must be mapped to the representation expected by the pretrained policy. Third, temporal evidence must augment the current observation, which remains essential for interpreting the present scene. Together, these requirements call for a selective residual pathway with explicit correspondence and a direct current-observation path.

\begin{figure*}[!t]
\centering
\includegraphics[width=\textwidth]{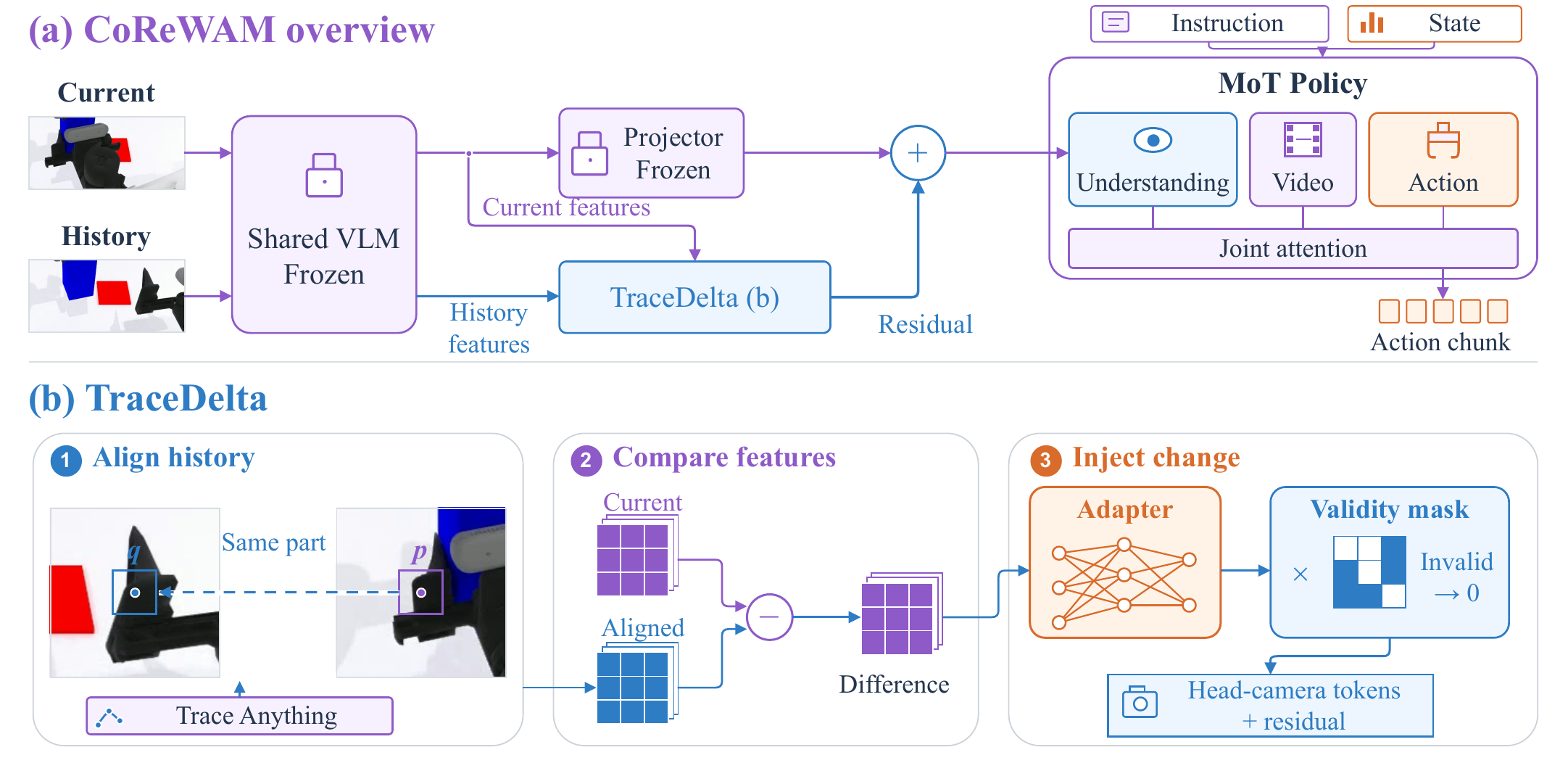}
\caption{\method\ retains current-observation conditioning while adding correspondence-aligned visual change. (a) A shared frozen vision-language model encodes current and historical observations. Current features pass through the frozen projector; TraceDelta uses both feature streams, and its residual is added to the projected current tokens. Instruction and state condition the Motus understanding, video, and action experts, whose joint attention generates an action chunk. (b) Frozen Trace Anything maps current point $p$ to historical point $q$ and transports its feature. TraceDelta subtracts this aligned feature from the current feature, adapts the signed difference, masks invalid entries to zero, and updates only head-camera tokens. Images and correspondences are schematic.}
\label{fig:overview}
\end{figure*}

We propose \method, whose TraceDelta interface realizes this pathway through three steps (Fig.~\ref{fig:overview}). TraceDelta first uses a frozen Trace Anything model~\cite{liu2025traceanything} to transport historical visual features to current token locations. It then computes a signed difference between current features and their aligned historical counterparts. Finally, a lightweight adapter maps this difference into the policy interface, where correspondence validity gates the residual before it is added to head-camera tokens after the frozen visual-language projector. The original Motus joint-attention pathway connects the conditioned visual representation to action generation. In this way, \method\ makes correspondence-aligned scene evolution available alongside the current observation without introducing a separate trajectory-to-action fusion module.

Our Motus implementation keeps the pretrained visual encoder and expert weights fixed and optimizes the temporal adapter, action-side low-rank updates~\cite{hu2021lora}, and the existing action decoder, totaling approximately 1.59 million trainable parameters. Across 50 clean RoboTwin 2.0 tasks~\cite{chen2025robotwin2}, experiments show improved spatial consistency and success rates for CoReWAM. We report skill-level results and temporal-component controls. Integrating TraceDelta into a StarVLA-based policy raises clean success from 58.10\% to 67.62\%, providing evidence of transferability beyond Motus. Section~\ref{sec:experiments} further examines robustness and physical deployment on Astribot S1. Our contributions are:
\begin{enumerate}
\item We introduce TraceDelta, a temporal-conditioning formulation that organizes visual comparison around corresponding scene content by transporting historical pretrained features before computing their signed difference (Secs.~\ref{sec:transport}--\ref{sec:differencing}).
\item We develop \method, a parameter-efficient Motus-based world-action model that integrates TraceDelta through a validity-gated residual pathway, making aligned history available for action generation while retaining current-observation conditioning and the pretrained expert architecture (Sec.~\ref{sec:conditioning}).
\item We evaluate \method\ across 50 RoboTwin 2.0 tasks spanning diverse manipulation skills. Under a 5,000-update adaptation budget, it reaches 92.22\% clean success, 3.56 points above Motus and 2.06 points above Raw Delta. On Astribot S1, it achieves 28/30 successes on an in-distribution task and 25/30 on an OOD task (Secs.~\ref{sec:comparison} and~\ref{sec:real}).
\end{enumerate}

\section{Related Work}
\label{sec:related_work}

\subsection{World-Action Models and Predictive Representations}
World-action models connect visual prediction with action generation to learn representations of how actions affect the environment~\cite{du2023unipi,wu2023gr1}. Motus~\cite{bi2025motus} integrates visual-language understanding, video generation, and action prediction through interacting transformer experts. Within this predictive framework, Fast-WAM~\cite{yuan2026fastwam} distinguishes the benefits of video co-training from the need for explicit future generation at inference, while DreamWAM~\cite{yuan2026dreamwam} extends future modeling beyond RGB to motion, geometry, and semantic representations. These studies investigate both how predictive learning supports control and what information a model should predict~\cite{assran2025vjepa2}. More recently, StageWAM~\cite{liu2026stagewam} distinguishes local visual evolution from task-stage progress, using a goal-conditioned JEPA predictor to estimate a next-stage representation that conditions the Motus video stream.

Our work addresses a complementary question: how should a policy relate its current observation to its observed history? Rather than adding another future prediction target, \method\ constructs correspondence-aligned temporal residuals from available observations. These residuals supplement visual conditioning while retaining the backbone's native prediction objectives.

\subsection{Motion and Trajectory Representations for Control}
Motion representations provide a structured connection between visual observations and robot behavior. Any-point Trajectory Modeling~\cite{wen2024atm} predicts future point trajectories and uses these predictions to guide policy learning. Moving from explicit trajectories to learned motion representations, CoMo~\cite{yang2026como} combines temporal feature differences with contrastive learning to extract continuous latent motion, which supplies pseudo-action labels for policy co-training. Motion can also enter a policy through historical evidence: MotionVLA~\cite{yuan2026motionvla} uses current visual tokens to retrieve historical motion features from a trajectory extractor and fuses the retrieved features into a vision-language-action model. Alternatively, 4D-WAM~\cite{yang2026fourDwam} transfers trajectory structure through auxiliary supervision, aligning temporal feature changes and source-conditioned destination relations with a frozen trajectory-field teacher.

TraceDelta assigns correspondence a different role: it acts as a transport operator over historical VLM features before temporal differencing. Unlike retrieving motion features or supervising representations with trajectory targets, this operation determines which historical visual content is compared with the present. The resulting policy conditioning expresses change within the pretrained visual-feature space.

\subsection{Correspondence and Aligned Feature Comparison}
Temporal comparison requires associating scene content across observations, since a fixed image location need not depict the same content over time~\cite{harley2022pips,doersch2023tapir}. Correspondence methods establish these associations at different scales~\cite{wang2023dust3r,wang2025vggt}: RAFT~\cite{teed2020raft} estimates dense optical flow between image pairs, CoTracker3~\cite{karaev2024cotracker3} tracks points across video, and Trace Anything~\cite{liu2025traceanything} represents pixels through continuous 3D trajectories in a shared coordinate system. Beyond estimating motion, correspondence can organize feature comparison~\cite{zhu2017dff,zhu2017fgfa}. CYWS-3D~\cite{sachdeva2023cyws3d} registers visual features across views before differencing them, then uses visibility masks and a detection head to localize scene changes.

We bring this alignment-before-comparison principle to robot-policy conditioning. Trajectory-derived correspondence transports historical features to the current observation's spatial support, after which a signed difference is mapped through a validity-gated adapter into the existing visual-token interface. Unlike change-localization methods, the output is an additive conditioning residual for action generation. The direct current-observation pathway remains intact, allowing temporal evidence to supplement the representation of the present scene.

\section{Method}
\label{sec:method}

\method\ makes correspondence-aligned short-term visual change
available for action generation while retaining
current-observation conditioning. Its TraceDelta interface
follows three steps (Fig.~\ref{fig:overview}(b)): \emph{align history} through
correspondence-based feature transport, \emph{compare features}
through signed temporal differencing, and \emph{inject change}
through validity-gated residual conditioning. We describe its
Motus instantiation~\cite{bi2025motus}, whose understanding, video-generation,
and action experts interact through joint attention. The
dimensions and token layout below refer to this implementation.

\subsection{Overview and Formulation}
\label{sec:overview}

At time $t$, the policy receives a current observation $O_t$,
robot state $s_t$, and instruction $\ell$, and produces an
action chunk $A_t$~\cite{zhao2023act,chi2023diffusionpolicy}. We additionally use a historical
observation $O_h$, where $h=t-s$ and $s$ is the observation
stride. The same frozen vision-language encoder $E_\phi$
processes both observations~\cite{qwen2025qwen3vl}:
\begin{equation}
    Z_\tau = E_\phi(O_\tau,\ell),
    \qquad \tau \in \{h,t\}.
    \label{eq:shared_encoding}
\end{equation}
Let $F_\tau \in \R^{H \times W \times D}$ denote the
spatial visual-feature grid extracted from $Z_\tau$, and let
$\Omega$ denote the head-camera token positions. The original
understanding input is $U_t=P(Z_t)$, where $P$ is the frozen
visual-language projector.

TraceDelta compares current and aligned historical features
in the encoder feature space, then maps their difference to
a residual in the projected interface. This residual is added
only at positions in $\Omega$, leaving the direct
current-observation pathway intact (Fig.~\ref{fig:overview}(a)).
Algorithm~\ref{alg:tracedelta} summarizes the causal forward
pass, where $\pi_\theta$ denotes the adapted Motus policy.

\begin{algorithm}[!t]
\caption{TraceDelta for Action Generation}
\label{alg:tracedelta}
\KwInput{Current observation $O_t$, optional history $O_h$, state $s_t$, and instruction $\ell$}
\KwOutput{Action chunk $A_t$}
$Z_t\leftarrow E_\phi(O_t,\ell)$; $U'_t\leftarrow P(Z_t)$\;
\If{historical observation $O_h$ is available}{
  $Z_h\leftarrow E_\phi(O_h,\ell)$\;
  Extract spatial feature grids $F_t,F_h$ from $Z_t,Z_h$\;
  Construct backward map $\mathcal{C}_{t\rightarrow h}$ and sampling-validity mask $m$ using frozen Trace Anything\;
  Initialize $r_i\leftarrow 0$ for all $i\in\Omega$\;
  \ForEach{head-camera token $i\in\Omega$}{
    \If{$m_i=1$}{
      $q_i\leftarrow\mathcal{C}_{t\rightarrow h}(p_i)$\;
      $\bar f_{h,i}\leftarrow\mathcal{W}(F_h,q_i)$ \tcp*[r]{Align history}
      $\Delta_i^{\mathrm{trace}}\leftarrow F_t(p_i)-\bar f_{h,i}$ \tcp*[r]{Compare features}
      $r_i\leftarrow\alpha m_i g_\psi(\Delta_i^{\mathrm{trace}})$ \tcp*[r]{Adapt + gate}
    }
  }
  $U'_t\leftarrow P(Z_t)+\operatorname{Scatter}_{\Omega}\!\left(\{r_i\}_{i\in\Omega}\right)$\;
}
$A_t\leftarrow\pi_\theta(U'_t,s_t,\ell)$\;
\Return{$A_t$}\;
\end{algorithm}

\subsection{Trajectory-Grounded Feature Transport}
\label{sec:transport}

Feature transport establishes which historical content should
be compared with each current visual token. Frozen Trace
Anything~\cite{liu2025traceanything} estimates dense pixel trajectories in a shared
3D coordinate system. For a current token center $p_i$, we
evaluate its associated trajectory at historical time $h$ and
match that point to the historical dense 3D field by
nearest-neighbor association. The matched historical pixel is
converted to a visual-feature sampling coordinate, accounting
for image resizing, cropping, and the camera's placement in
the packed visual grid. This defines a backward map
$\mathcal{C}_{t\rightarrow h}$:
\begin{equation}
    q_i = \mathcal{C}_{t\rightarrow h}(p_i),
    \qquad
    \bar f_{h,i} = \mathcal{W}(F_h,q_i),
    \quad i \in \Omega,
    \label{eq:feature_transport}
\end{equation}
where $\mathcal{W}$ denotes bilinear sampling~\cite{jaderberg2015stn}. The
transported feature $\bar f_{h,i}$ represents the estimated
historical counterpart of the current region, rather than
content at the same historical image coordinate.

A binary mask $m_i$ records whether the sampling
correspondence is usable: it must be finite and fall within
the designated camera region. Invalid coordinates are
excluded from residual injection. This mask specifies
admissible sampling support, not a learned semantic score.
Correspondence thus determines where history is sampled;
neither trajectory coordinates nor trajectory-encoder
features are appended as additional policy tokens.

\subsection{Signed Temporal Differencing}
\label{sec:differencing}

After establishing correspondence, TraceDelta measures
feature change between current regions and their historical
counterparts. For comparison, a fixed-grid temporal residual,
which we call Raw Delta, is
\begin{equation}
    \Delta_i^{\mathrm{raw}}
    = F_t(p_i)-F_h(p_i).
    \label{eq:raw_delta}
\end{equation}
When objects or the camera move, the same grid location may
describe different scene content across observations. Its
difference then reflects changing spatial occupancy as well
as changes in visual features. TraceDelta instead compares
each current feature with its transported historical
counterpart:
\begin{equation}
    \begin{aligned}
        \Delta_i^{\mathrm{trace}}
        &= F_t(p_i)-\bar f_{h,i} \\
        &= F_t(p_i)
        - \mathcal{W}\!\left(F_h,\mathcal{C}_{t\rightarrow h}(p_i)\right).
    \end{aligned}
    \label{eq:trace_delta}
\end{equation}

Both operands lie in the same frozen feature space.
Retaining the sign preserves the direction of change within
this space rather than collapsing it into a magnitude.
Correspondence determines which historical region is
compared with the current region; the difference measures
feature change after this association, not displacement
itself. An identity correspondence reduces
Eq.~\eqref{eq:trace_delta} to Raw Delta and does not, in
general, make the difference zero.

\subsection{Validity-Gated Visual Conditioning}
\label{sec:conditioning}

To make aligned feature change available for action
generation, a lightweight adapter maps it to the Motus
understanding dimension:
\begin{equation}
    g_\psi(\Delta)
    =
    W_2\,\operatorname{GELU}\!\left(
        W_1\,\operatorname{LN}(\Delta)+b_1
    \right)+b_2.
    \label{eq:temporal_adapter}
\end{equation}
Our implementation uses dimensions
$2048 \rightarrow 256 \rightarrow 512$.
A learned scalar $\alpha$ scales the residual, and sampling
validity gates it after adaptation:
\begin{equation}
    r_i
    =
    \alpha m_i
    g_\psi\!\left(\Delta_i^{\mathrm{trace}}\right).
    \label{eq:gated_residual}
\end{equation}
Post-adapter masking ensures that invalid locations
contribute zero even when the adapter has nonzero biases.
The residual is then scattered into the native head-camera
positions:
\begin{equation}
    U'_t
    =
    P(Z_t)
    +
    \operatorname{Scatter}_{\Omega}
    \left(\{r_i\}_{i\in\Omega}\right).
    \label{eq:visual_conditioning}
\end{equation}

In the evaluated head-camera configuration, $\Omega$
comprises 60 tokens arranged in a $6 \times 10$ grid, at
indices 10--69 within the packed visual-token sequence.
Other positions receive no direct residual. The
current-observation path $P(Z_t)$ remains intact, while
the existing joint-attention computation propagates the
augmented visual conditioning to the action expert.
This lets the policy use observed feature changes alongside
the present scene without a separate trajectory-to-action
fusion module.

\subsection{Training and Causal Inference}
\label{sec:training}

During adaptation, the pretrained backbone weights remain
fixed. We optimize the temporal adapter, action-side low-rank
updates~\cite{hu2021lora}, and the existing action decoder, totaling
approximately 1.59 million trainable parameters, or
$0.02\%$ of the policy. The RoboTwin adaptation uses
$1/192$ as many training-sample presentations as 4D-WAM's
reported FastWAM-Joint schedule~\cite{yang2026fourDwam}, where presentations
are measured as global batch size $\times$ update count.
This ratio concerns the additional adaptation budget,
rather than end-to-end computational cost.

At inference, frozen Trace Anything estimates
correspondence online using only current and past
observations. Training and inference share the same VLM
preprocessing and head-camera token mapping. When history
is unavailable, the temporal branch is bypassed and the
policy uses $P(Z_t)$ alone for visual conditioning.

\section{Experiments}
\label{sec:experiments}

We evaluate \method\ on RoboTwin, examine transferability to a StarVLA-based policy, and report temporal-component controls and physical deployment on Astribot S1.

\subsection{Experimental Setup}
\label{sec:setup}
\noindent\textbf{Base model and baselines.}
We build \method\ on Motus~\cite{bi2025motus}, whose shared attention connects visual understanding, video generation, and action prediction. We choose it because its visual-conditioning interface lets us test aligned history while preserving the expert architecture. Motus already learns motion priors through optical-flow-based pretraining, allowing us to test the additional value of explicit correspondence. We compare with $\pi_{0.5}$~\cite{physicalintelligence2025pi05}, X-VLA~\cite{zheng2025xvla}, Fast-WAM and FastWAM-Joint~\cite{yuan2026fastwam}, Motus, and 4D-WAM~\cite{yang2026fourDwam}. 4D-WAM builds on FastWAM-Joint.

\noindent\textbf{Data and training.}
Our primary benchmark is RoboTwin 2.0~\cite{chen2025robotwin2}, with 50 manipulation tasks. Additional adaptation uses only clean data: 2,386 episodes provide 19,088 samples, with episode-disjoint held-out data for checkpoint selection. We train for 5,000 updates on eight NVIDIA H100 GPUs, using batch size four per GPU (global batch 32). The backbone stays frozen. The temporal adapter adds 0.66M parameters; with action-side low-rank updates, newly introduced parameters total 1.58M. Including the existing action decoder, 1.59M parameters are optimized, approximately 0.02\% of the policy.

\noindent\textbf{Evaluation.}
We report success rates (SR), averaged equally across all 50 tasks, and baseline-relative gains $\Delta$ in percentage points (pp) on each split. Local controls share the starting policy, adaptation data, and 5,000-update budget. Published references retain their source protocols. For TraceDelta, clean and randomized results follow an aggregation protocol across two seeds.

\subsection{Performance on RoboTwin 2.0}
\label{sec:comparison}
We evaluate \method\ across all 50 RoboTwin 2.0 tasks under clean and randomized settings. Table~\ref{tab:robotwin} summarizes the overall success rates alongside representative vision-language-action and world-action policies. These references provide context for the benchmark; our component study examines the contribution of the temporal interface within the same policy architecture.

\noindent\textbf{Overall results.}
\method\ achieves \traceClean\% clean success in Table~\ref{tab:robotwin_tasks}, and \traceRand\% randomized success, with both results aggregated across two seeds. Relative to the base policy's reported scores, the gains are 3.56 and 2.58 pp, respectively. The temporal interface retains the current-observation pathway and adds aligned historical changes without extending the policy's visual-token sequence. Section~\ref{sec:ablation} examines how action-side adaptation, current-feature conditioning, and temporal differencing contribute to these results.

\begin{table}[!htbp]
\centering
\caption{RoboTwin 2.0 success (\%) and reported gains (pp). Bold/underline mark the highest/second-highest SR in each column. Shading identifies our method. $\Delta_C$/$\Delta_R$ are relative to Motus, except for 4D-WAM, which uses FastWAM-Joint. Protocols follow Section~\ref{sec:setup}.}
\label{tab:robotwin}
\small
\setlength{\tabcolsep}{4pt}
\renewcommand{\arraystretch}{1.12}
\begin{tabularx}{\columnwidth}{@{}Xrrrr@{}}
\toprule
Method & Clean & Rand. & $\Delta_C$ & $\Delta_R$\\
\midrule
$\pi_{0.5}$ & 82.74 & 76.76 & --- & ---\\
X-VLA & 72.88 & 72.84 & --- & ---\\
Track4Action~\cite{wang2026track4action} & 80.44 & 81.48 & --- & ---\\
Fast-WAM & 91.88 & \textbf{91.78} & --- & ---\\
FastWAM-Joint & 90.84 & 90.32 & --- & ---\\
Motus & 88.66 & 87.02 & --- & ---\\
4D-WAM & \underline{91.92} & \underline{90.76} & +1.08 & +0.44\\
\midrule
Raw Delta & \rawClean & \rawRand & \rawCleanDelta & \rawRandDelta\\
\rowcolor{corepurplefill} \textbf{\method} & \textbf{\traceClean} & \traceRand & \gain{\traceCleanDelta} & \gain{\traceRandDelta}\\
\bottomrule
\end{tabularx}
\end{table}

\noindent\textbf{Task-level results.}
Table~\ref{tab:robotwin_tasks} expands the aggregate evaluation with 18 tasks showing gains of at least 5 pp over the base policy. The final row reports the mean over all 50 tasks. The selected tasks cover object transfer, handover, placement, stacking, and tool use, providing a closer view of performance across different manipulation behaviors.

The largest gains within this subset occur in Move Can Pot (+50 pp), Scan Object (+19 pp), and Handover Mic (+16 pp). Placement and packing also improve: Put Bottles Dustbin gains 13 pp, while Place A2B Left and Place Bread Skillet each gain 10 pp. These results complement the full-benchmark summary.

\begin{figure*}[!t]
\centering
\includegraphics[width=\textwidth]{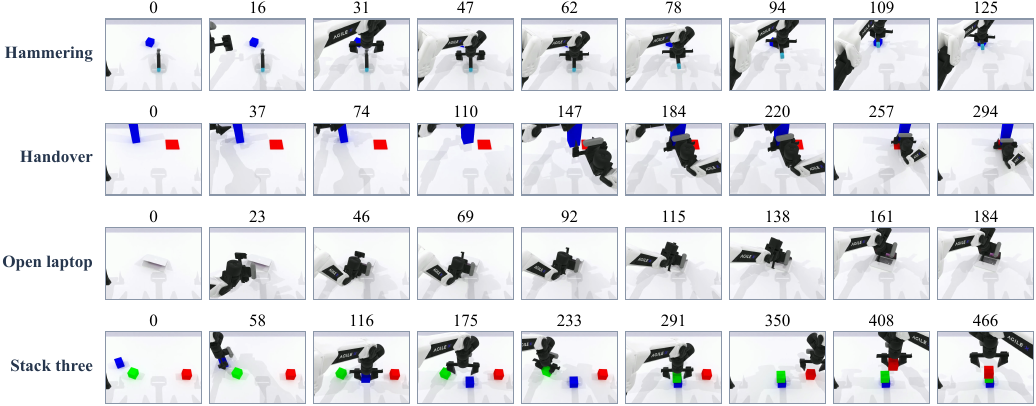}
\caption{Four RoboTwin examples show how task-relevant scene content evolves during manipulation. From top to bottom, the rows depict hammering, object handover, opening a laptop, and stacking three objects. Each row contains nine uniformly sampled head-camera frames; the numbers above them are original frame indices. The strips visualize the temporal evidence available to TraceDelta, which compares corresponding content within a short local history before conditioning the next action chunk.}
\label{fig:motion_sequence}
\end{figure*}

Figure~\ref{fig:motion_sequence} illustrates histories from hammering, handover, articulation, and stacking. These sequences show the changing visual content available to TraceDelta; the component controls below examine how the policy incorporates this history.

\begin{table*}[!t]
\centering
\caption{Task-level clean success rates (\%) on RoboTwin 2.0. The 18 tasks with gains of at least 5 pp over the base policy are ordered by $\Delta$. Bold/underline mark the highest/second-highest displayed score, including ties. Shaded columns show \method\ and its gain over Motus. TraceDelta clean results are aggregated across two seeds following Section~\ref{sec:setup}; the last row equally weights all 50 tasks.}
\label{tab:robotwin_tasks}
\small
\setlength{\tabcolsep}{5pt}
\begin{tabular*}{\textwidth}{@{\extracolsep{\fill}}lrrrr>{\columncolor{corepurplefill}}r>{\columncolor{corepurplefill}}r@{}}
\toprule
Selected task & $\pi_{0.5}$ & X-VLA & FastWAM-Joint & Motus & \method & $\Delta$\\
\midrule
Move Can Pot & 51 & \underline{89} & \textbf{97} & 34 & 84 & \gain{+50}\\
Scan Object & 72 & 14 & \textbf{92} & 67 & \underline{86} & \gain{+19}\\
Handover Mic & \underline{98} & 0 & \textbf{100} & 78 & 94 & \gain{+16}\\
Put Bottles Dustbin & 84 & 74 & \underline{93} & 81 & \textbf{94} & \gain{+13}\\
Place A2B Left & 87 & 48 & \underline{96} & 88 & \textbf{98} & \gain{+10}\\
Place Bread Skillet & 85 & 77 & \underline{90} & 86 & \textbf{96} & \gain{+10}\\
Place Mouse Pad & 60 & 70 & \textbf{96} & 66 & \underline{76} & \gain{+10}\\
Rotate QR Code & 89 & 34 & \underline{91} & 89 & \textbf{98} & \gain{+9}\\
Stack Bowls Three & 77 & 76 & \underline{84} & 79 & \textbf{88} & \gain{+9}\\
Place Object Basket & 80 & 44 & \underline{86} & 81 & \textbf{89} & \gain{+8}\\
Move Pill Bottle to Pad & 84 & 73 & \underline{99} & 93 & \textbf{100} & \gain{+7}\\
Place Fan & 87 & 80 & \textbf{99} & 91 & \underline{98} & \gain{+7}\\
Place Object Scale & 86 & 52 & \textbf{96} & 88 & \underline{94} & \gain{+6}\\
Press Stapler & 87 & 92 & 52 & \underline{93} & \textbf{99} & \gain{+6}\\
Dump Bin into Big Bin & 92 & 79 & \underline{95} & \underline{95} & \textbf{100} & \gain{+5}\\
Place Can Basket & 62 & 49 & 50 & \underline{81} & \textbf{86} & \gain{+5}\\
Place Phone Stand & 81 & 88 & \textbf{100} & 87 & \underline{92} & \gain{+5}\\
Stamp Seal & 79 & 76 & \underline{96} & 93 & \textbf{98} & \gain{+5}\\
\midrule
\textbf{All tasks (50)} & 82.74 & 72.88 & \underline{90.84} & 88.66 & \textbf{92.22} & \gain{+3.56}\\
\bottomrule
\end{tabular*}

\par\smallskip
\begin{minipage}{\textwidth}\footnotesize
References are from Fast-WAM Table 3~\cite{yuan2026fastwam} for $\pi_{0.5}$/FastWAM-Joint and StageWAM Table 6~\cite{liu2026stagewam} for X-VLA/Motus. The last row reports all 50 tasks.
\end{minipage}
\end{table*}

\subsection{Transferability Evaluation}
\label{sec:transferability}
\label{sec:clean_transfer}
\noindent\textbf{Transfer across policy architectures.}
A temporal interface is more broadly useful if its benefit extends beyond the architecture on which it was developed. We therefore integrate TraceDelta into a StarVLA-based policy~\cite{starvla2026}, testing whether correspondence-aligned history also improves a second visual/action interface. This evaluation transfers the temporal-conditioning design: align historical visual features, compute their difference from current features, and make that residual available to the policy.

\noindent\textbf{Matched adaptation protocol.}
We compare the StarVLA baseline with its TraceDelta-augmented counterpart using the same clean RoboTwin data, adaptation budget, and clean-only checkpoint-selection criterion. Both variants are adapted for this evaluation. The comparison thus tests the benefit of transferring the interface under a matched training budget, rather than reusing a Motus-trained checkpoint without adaptation.

\noindent\textbf{A 9.52-point gain beyond Motus.}
TraceDelta improves clean success from 58.10\% to 67.62\% (Table~\ref{tab:clean_transfer}), a gain of \textbf{9.52 percentage points}. This improvement shows that the value of aligned temporal conditioning extends to a second policy architecture. Together with the Motus results, it supports the transferability of the core operation---comparing corresponding scene content before conditioning action generation---across different visual/action interfaces.

\noindent\textbf{Transferability and distribution shift.}
We also evaluate the clean-trained variants under randomized conditions, where success changes from 10.60\% to 10.81\% (+0.21 pp). These two columns address different questions: the clean comparison establishes the benefit of the transferred interface, while randomized evaluation measures its behavior under distribution shift. The strongest transfer result is the 9.52-point clean improvement under matched adaptation. The 1.59M-parameter budget in Table~\ref{tab:parameter_budget} describes our Motus implementation; the StarVLA experiment establishes architectural transferability separately.

\begin{table}[!htbp]
\centering
\caption{Transferability to StarVLA under the same clean-data adaptation budget. SR is in \%; gains are relative to StarVLA.}
\label{tab:clean_transfer}
\small
\setlength{\tabcolsep}{4pt}
\renewcommand{\arraystretch}{1.15}
\begin{tabularx}{\columnwidth}{@{}Xrr@{}}
\toprule
Policy & Clean & Randomized\\
\midrule
StarVLA & 58.10 & 10.60\\
\rowcolor{corepurplefill} StarVLA + TraceDelta & \textbf{67.62} & \textbf{10.81}\\
\midrule
Gain (pp) & \gain{+9.52} & \gain{+0.21}\\
\bottomrule
\end{tabularx}
\end{table}

\begin{table*}[!t]
\centering
\caption{Component analysis on RoboTwin 2.0. All adapted variants update the action decoder; LoRA denotes additional action-side low-rank updates. Params: optimized parameters (M). SR is in \%; Avg. equally weights Clean and Randomized. Parentheses show the numerical gap to the complete model's reported Avg. (pp).}
\label{tab:ablation}
\small
\setlength{\tabcolsep}{5pt}
\renewcommand{\arraystretch}{1.18}
\begin{tabularx}{\textwidth}{@{}l*{4}{>{\centering\arraybackslash}X}rrrr@{}}
\toprule
& \multicolumn{4}{c}{Adaptation components} & & \multicolumn{3}{c}{Success rate (\%)}\\
\cmidrule(lr){2-5}\cmidrule(l){7-9}
Variant & LoRA & Adapter & History & Align & Params & Clean & Rand. & Avg.\\
\midrule
Motus (published) & \no & \no & \no & \no & --- & 88.66 & 87.02 & 87.84\ \gap{-3.07}\\
\midrule
LoRA-only & \yes & \no & \no & \no & 0.934 & 90.90 & 87.30 & 89.10\ \gap{-1.81}\\
Current-only & \yes & \yes & \no & \no & 1.594 & 91.34 & 87.82 & 89.58\ \gap{-1.33}\\
Raw Delta & \yes & \yes & \yes & \no & 1.594 & \rawClean & \rawRand & \rawAvg\ \gap{-2.00}\\
\midrule
\rowcolor{corepurplefill}
\textbf{TraceDelta (ours)} & \yes & \yes & \yes & \yes & 1.594 & \textbf{\traceClean} & \textbf{\traceRand} & \textbf{\traceAvg}\\
\bottomrule
\end{tabularx}
\par\smallskip
\begin{minipage}{\textwidth}\footnotesize
Component marks refer to the additional adaptation pathway, not the pretrained backbone.
\end{minipage}
\end{table*}

\subsection{Ablation Study}
\label{sec:ablation}
We examine three questions: what action-side adaptation contributes, whether an additional current-feature adapter is sufficient, and how temporal conditioning changes when historical features are aligned. Table~\ref{tab:ablation} makes the active components explicit; Table~\ref{tab:parameter_budget} reports the parameter budget of the complete model.

\noindent\textbf{Controlled adaptation setup.}
LoRA-only, Current-only, and Raw Delta use the same experimental configuration. Their validation-selected checkpoints are evaluated on all 50 RoboTwin 2.0 tasks under clean and randomized settings. All three update the action-side low-rank weights and action decoder. Current-only and Raw Delta also use the same visual-adapter architecture as TraceDelta, so their distinction is the adapter input rather than additional visual-adapter capacity.

\noindent\textbf{What each variant tests.}
\emph{LoRA-only} adapts the action pathway without injecting a visual residual. It provides the reference for evaluating the added visual interface. \emph{Current-only} supplies current head-camera features to the visual adapter but does not use history; it tests whether extra visual-adapter capacity alone accounts for performance. \emph{Raw Delta} supplies $F_t(p_i)-F_h(p_i)$ to that adapter, introducing temporal information at fixed grid locations. \emph{TraceDelta} transports history with $\mathcal{C}_{t\rightarrow h}$ before differencing, as in Eq.~\eqref{eq:trace_delta}. Its correspondence provider is frozen, so alignment does not add optimized parameters to the visual residual pathway.

\noindent\textbf{Visual conditioning beyond action adaptation.}
Current-only reaches 91.34\% clean and 87.82\% randomized success, compared with 90.90\% and 87.30\% for LoRA-only. The respective gains of 0.44 and 0.52 pp show that the lightweight visual interface contributes alongside action-side adaptation. This comparison motivates retaining Current-only as a capacity-matched reference for temporal conditioning.

\noindent\textbf{History and correspondence alignment.}
Raw Delta reaches 90.16\% clean and 87.66\% randomized success. Relative to Current-only, its effect depends on the task: the four illustrated manipulation types in Table~\ref{tab:ablation_tasks} show both positive and negative changes, while the full 50-task means differ by $-1.18$ and $-0.16$ pp on clean and randomized evaluation. Historical input and a useful temporal comparison are therefore distinct design choices.

The complete TraceDelta model reports 92.22\% clean and 89.60\% randomized success. These are 2.06 and 1.94 pp above Raw Delta, and 0.88 and 1.78 pp above Current-only. Together with the explicit component structure, these results motivate correspondence-aligned temporal conditioning: the adapter receives change between matched scene regions rather than a mixture of feature change and shifting grid occupancy.

\begin{table}[!htbp]
\centering
\caption{Task-level randomized controls (\%). The four tasks correspond to the manipulation types in Fig.~\ref{fig:motion_sequence}; the final row covers all 50 tasks. Bold marks the highest score in each row.}
\label{tab:ablation_tasks}
\footnotesize
\setlength{\tabcolsep}{3pt}
\renewcommand{\arraystretch}{1.13}
\begin{tabular*}{\columnwidth}{@{\extracolsep{\fill}}lrrr@{}}
\toprule
Task & LoRA-only & Current-only & Raw Delta\\
\midrule
Beat Block Hammer & 82 & 85 & \textbf{88}\\
Handover Block & 75 & 80 & \textbf{82}\\
Open Laptop & 86 & \textbf{92} & 90\\
Stack Blocks Three & 68 & 71 & \textbf{73}\\
\midrule
All tasks (50) & 87.30 & \textbf{87.82} & 87.66\\
\bottomrule
\end{tabular*}
\end{table}

\noindent\textbf{Parameter efficiency.}
Table~\ref{tab:parameter_budget} separates the temporal interface from action-side adaptation. The adapter and its residual scale account for 660,225 parameters. Action LoRA adds 917,504, and the existing action decoder contributes 16,398 optimized parameters. The total is 1,594,127, approximately 0.02\% of the policy. Current-only, Raw Delta, and TraceDelta retain the same adapter dimensions; correspondence alignment changes historical sampling without increasing the trainable adapter or the number of policy input tokens. Frozen-backbone and frozen-correspondence parameters are excluded from this optimization budget.

\begin{table}[!htbp]
\centering
\caption{Optimized parameter budget for the Motus-based model. Newly introduced parameters exclude the existing action decoder. Only approximately 0.02\% of the Motus-based policy parameters are optimized; the pretrained backbone and correspondence estimator remain frozen.}
\label{tab:parameter_budget}
\small
\setlength{\tabcolsep}{4pt}
\renewcommand{\arraystretch}{1.13}
\begin{tabularx}{\columnwidth}{@{}Xr@{}}
\toprule
Component & Parameters\\
\midrule
Temporal adapter and residual scale & 660,225\\
Action-side LoRA & 917,504\\
Existing action decoder & 16,398\\
\midrule
Newly introduced & 1,577,729\\
\rowcolor{corepurplefill} \textbf{Total optimized} & \textbf{1,594,127}\\
\rowcolor{corepurplefill} \textbf{Share of policy parameters} & \textbf{$\approx$0.02\%}\\
\bottomrule
\end{tabularx}
\end{table}

\begin{figure*}[!t]
\centering
\includegraphics[width=\textwidth]{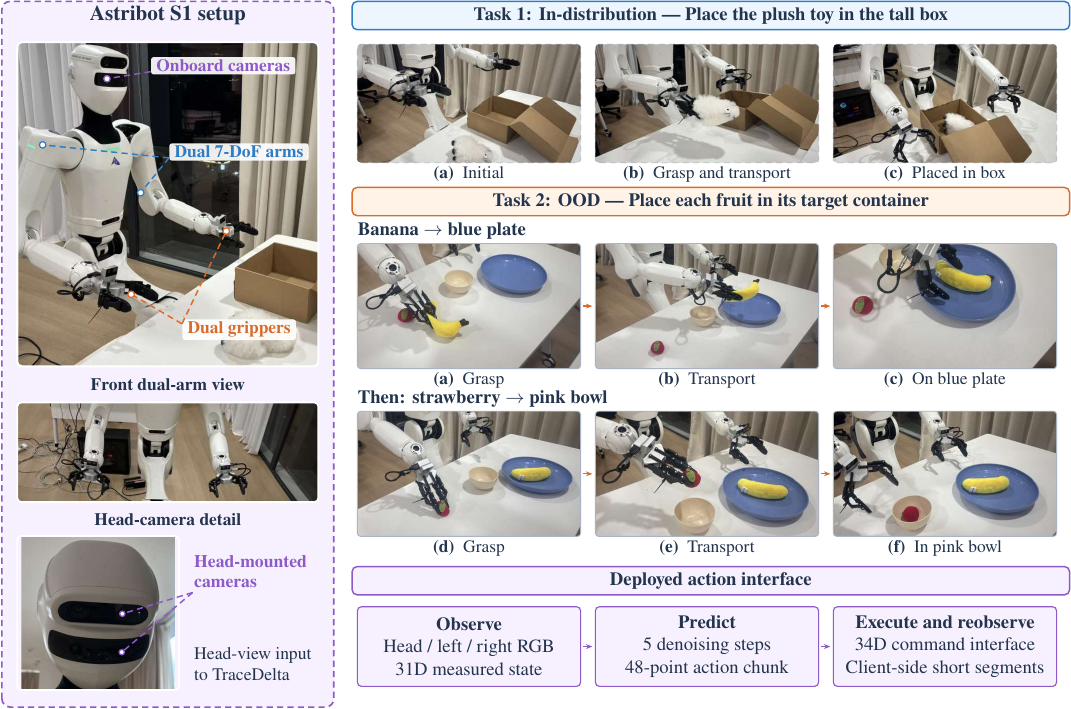}
\caption{Physical deployment on Astribot S1. Left: onboard cameras, dual arms and grippers, and TraceDelta's head-view input. Right: plush-toy placement and an OOD sequence placing a banana on a blue plate, then a strawberry in a pink bowl. Bottom: the policy observes three RGB views and robot state, predicts a 48-step action chunk, executes a short segment, and reobserves.}
\label{fig:real_robot_overview}
\end{figure*}

\subsection{Real-World Deployment}
\label{sec:real}
We fine-tune \method\ for Astribot S1 while keeping WAN2.2 and Qwen3-VL frozen, and deploy the validation-selected checkpoint on the physical robot (Fig.~\ref{fig:real_robot_overview}). The policy observes head, left, and right RGB views together with a 31-dimensional measured robot state and predicts 48-step Cartesian action sequences. Before execution, denormalization and observed-chassis-pose augmentation convert the 31-dimensional predictions into the robot's 34-dimensional command interface.

\noindent\textbf{Physical tasks and success.}
Each physical task uses 30 trials, with success defined as completing the full instruction (Table~\ref{tab:real_robot}). \method\ completes in-distribution plush-toy placement in 28 trials (93.3\%) and the OOD sequence of placing a banana on a blue plate followed by a strawberry in a pink bowl in 25 trials (83.3\%). Motus achieves 20/30 (66.7\%) and 18/30 (60.0\%) successes on these tasks, respectively.

\begin{table}[!htbp]
\centering
\caption{Real-world success on Astribot S1 over 30 trials per task. Entries show successes/trials (SR, \%). ID denotes in-distribution.}
\label{tab:real_robot}
\small
\setlength{\tabcolsep}{4pt}
\begin{tabularx}{\columnwidth}{@{}Xlrr@{}}
\toprule
Task & Split & Motus & \method\\
\midrule
Plush-toy placement & ID & 20/30 (66.7) & \textbf{28/30 (93.3)}\\
Fruit placement & OOD & 18/30 (60.0) & \textbf{25/30 (83.3)}\\
\bottomrule
\end{tabularx}
\end{table}

\noindent\textbf{Inference setup.}
At inference, we use five flow-denoising steps to generate 48-step action sequences corresponding to a 1.6\,s horizon. Commands are executed at 30\,Hz in short segments, with new observations acquired between segments. Offline profiling after warm-up
indicates a typical model-inference latency of approximately 510\,ms, excluding network communication and robot execution.

\FloatBarrier
\section{Conclusion}
\label{sec:conclusion}
We presented \method, a parameter-efficient world-action
model that uses correspondence-aligned visual changes for
action generation. Its TraceDelta module aligns historical
features before comparison and integrates the resulting
changes through a lightweight residual interface, while
retaining current-observation conditioning. With the
pretrained backbone frozen, \method\ optimizes only a
compact set of temporal and action-side parameters.
This modular design preserves the existing policy
architecture, facilitating adaptation to new manipulation
settings without full-model fine-tuning. Our experiments
show competitive performance across diverse manipulation
tasks, while deployment on Astribot S1 demonstrates
applicability to physical robot control.

A remaining limitation is the inference overhead introduced
by online correspondence estimation and temporal feature
processing. Future work will explore lightweight
correspondence estimation and feature reuse to reduce
latency and improve closed-loop responsiveness.

\bibliographystyle{IEEEtran}
\bibliography{references_trace_student}
\end{document}